\documentclass{article} % For LaTeX2e
\usepackage{iclr2021_conference,times}

\usepackage{hyperref}
\usepackage{booktabs}
\usepackage{url}
\usepackage[ruled]{algorithm2e}
\usepackage{algpseudocode}
\usepackage{mathtools}
\usepackage{verbatim}
\usepackage{subcaption}
\usepackage{caption}
\title{Prior-Independent Auctions for the Demand Side of Federated Learning}
\SetKwFunction{Punish}{Punish}
\SetKwFunction{Thresh}{Thresh}
\SetKwFunction{Score}{Score}
\SetKwProg{Fn}{Function}{:}{}

\SetKwFunction{Aggr}{Aggr}
\SetKwFunction{Train}{Train}
\SetKwFunction{FedEval}{FedEval}
\SetKwFunction{aggr}{Aggr}
\SetKwInOut{Input}{Input}
\iclrfinaltrue
\usepackage{amsthm}
\newtheorem{thm}{Theorem}

\newtheorem{assumption}[thm]{Assumption}
\author{Andreas~Haupt\\
LIDS\\
MIT\\
Cambridge, MA 02446, USA \\
\texttt{haupt@mit.edu} \\
\And
Vaikkunth Mugunthan \\
CSAIL \\
MIT \\
Cambridge, MA, 02446 \\
\texttt{vaik@mit.edu}
}
\begin{document}
\maketitle
\begin{abstract}
   Federated learning (FL) is a paradigm that allows distributed clients to learn a shared machine learning model without sharing their sensitive training data. While largely decentralized, FL requires resources to fund a central orchestrator or to reimburse contributors of datasets to incentivize participation. Inspired by insights from prior-independent auction design, we propose a mechanism, FIPIA (Federated Incentive Payments via Prior-Independent Auctions), to collect monetary contributions from self-interested clients. The mechanism operates in the semi-honest trust model and works even if clients have a heterogeneous interest in receiving high-quality models, and the server does not know the clients' level of interest. We run experiments on the MNIST, FashionMNIST, and CIFAR-10 datasets  to test clients' model quality under FIPIA and FIPIA's incentive properties.
\end{abstract}
\section{Introduction}
Federated learning (FL) \citep{bonawitz2019towards, mcmahan2017communication} is a mechanism that allows decentralized clients to collaboratively learn a machine learning model while preserving individual clients' privacy. That is, clients use datasets that share the same feature space but differ in sample space. However, without incentives, participating clients may provide obsolete information or decide to opt out of participation. Also, clients can submit inferior models yet end up receiving updated high-quality models. Hence, causing a free-rider problem \citep{Feldman2005OvercomingSystems}. 

There are numerous system designs \citep{Chen,Jiao2019,Li2019,yu2020fairness} that target this problem by having systems that reimburse clients that submit high-quality models. Such systems hence design incentives for the supply of models to FL. 

Correspondingly, we consider the clients receiving models and their incentives, the \emph{demand side of  FL}. We assume (an assumption to be relaxed in future work) that clients submit models trained on their entire dataset---with no need to incentivize them to do so. Under this assumption, we design a system to collect resources from clients.

To do so, we adapt auction designs to allocate freely-replicable (digital) goods \citep{hartline2013mechanism} to the FL setting and propose a system that works without central-entity evaluation. Our proposed system shares higher quality models with clients that pay more, as we demonstrate using the MNIST \citep{lecun1998mnist}, FashionMNIST \citep{xiao2017fashion}, and CIFAR-10 \citep{krizhevsky2009learning} datasets. We show experimentally that non-truthful bidding is only limitedly profitable.

\section{Literature Review}
First, our model connects to models incentivizing parity in test accuracy in FL. \citep{Li2019} proposes a variant of a global objective criterion to induce parity in test accuracy between clients holding non-i.i.d data. While this literature also studies the supply side, i.e., who receives models, the paper implicitly assumes that the interest in high-quality models is homogeneous across clients and tries to induce parity without collecting any monetary contributions.

Our approach also connects to the literature on incentive design for FL \citep{Chen,Jiao2019,yu2020fairness}. These papers propose systems that distribute a fixed \emph{incentive budget} to clients. Participating clients are promised payments depending on their model quality. These payments incentivize clients to \emph{contribute} high-quality models, but do not generate resources from clients that \emph{receive} models. Closest to our paper, \citep{Chen} studies how to incentivize clients to submit their full datasets to an FL system. Crucially, in their paper, the clients' private information is given by their maximal dataset size only, their interest in models they receive is assumed to be known to the system. One of the main properties of FIPIA is that it works without such knowledge.

This paper uses tools from the honest-but-curious or semi-honest trust model \citep{beaver1991foundations}. Our federated evaluation method is a variant of \citep{Mugunthan2020a}, which operates in the same trust model.

Finally, we use an incentive design, the circuit auction, from prior-independent mechanism design \citep{hartline2013mechanism}.

\section{Method}
We consider a horizontal FL setting. There are $k$ clients $i \in [k], k \coloneqq \{1, 2, \dots, k\}$ that can evaluate a model $w$ on their test sets to get a measure of quality such as accuracy, \Score{$i, w$}. In usual FL settings,  all clients receive the same models $w$. In our approach, we relax this assumption. That is, each client $i$ receives model $w_i$ after every federated round. To determine who will get which model, we also introduce payments $t_i, i=1, 2, \dots, k$, which clients pay to the server. These payments can be re-distributed amongst clients at a later stage using an existing incentive mechanism \citep{Chen,Jiao2019,yu2020fairness}.
\begin{assumption}\label{ass:utility}
Clients maximize a linear combination of payments they make and the quality of the model they receive,
\begin{equation}
\theta_i \text{\Score{$i, w_i$}} - t_i.\label{eq:util}
\end{equation}
$\theta_i$ represent client $i$'s valuation per unit of score. The valuations $\theta_i$ are not known to the server.
\end{assumption}
For example, Assumption\autoref{ass:utility} holds when clients incur a constant cost for each misclassified data point in their test set in each federated round. Implicit in this assumption is that clients are uninterested in other clients' model qualities, i.e., there are no externalities between clients.
\begin{assumption}\label{ass:data}
Clients submit models trained on their entire dataset.
\end{assumption}
This assumption mirrors our focus on the \emph{demand side} of FL. Our system leaves incentivizing high-quality models to other incentive mechanisms.
\begin{assumption}
The datasets used by clients are drawn from the same distribution (i.i.d. data).
\end{assumption}
An extension of the proposed system to the non-i.i.d. setting will involve additional complexities, compare \citep{hsu2019measuring}. We leave this for further work.

\subsection{Insufficiency of Average Accuracy Maximization}
A standard FL system wishes to maximize the average model quality amongst all clients, i.e., it maximizes $ \max_w \sum_{i=1}^k \lambda_i \text{\Score{$i,w$}}$, where $\lambda_i \ge 0$, denotes the relative importance of client $i$. In the notation introduced above, this corresponds to $w_i = w$, for $i=1, 2, \dots, k$. 

Assuming that a central orchestrator or engineer for the model needs to be funded, a classical FL system might not raise a sufficient amount of money to make the system sustainable. For example, if funding is done through voluntary monetary contributions by clients, a free-rider problem arises \citep[Section 11.C]{mas1995microeconomic}, and systems will likely be under-funded. Interestingly, the celebrated Vickrey-Clarke-Groves mechanism \citep{clarke1971multipart} fails to collect \emph{any} funds when applied on the demand side of FL. In this mechanism, each client that receives a model pays the system an amount corresponding to the negative effect it has on other clients. The receiving client pays the difference in valuation, according to \eqref{eq:util}, for all other clients when it participates in contrast to when it does not participate. As distributing the same model to an additional client does not have any effect on other clients in our model, \emph{clients pay nothing}. Hence, a system that collects funds in environments with costly orchestration is needed.
\subsection{Federated Incentive Payments via Prior-Independent Auctions (FIPIA)}
Our main contribution is a system for collecting payments in the semi-honest trust model, presented in \autoref{algo:main}. Our system is inspired by auction formats that guarantee (up to a multiplicative constant) maximal revenue in a wide range of theoretical setting satisfying Assumptions\autoref{ass:utility} and \autoref{ass:data} \citep{hartline2013mechanism,goldberg2001competitive}. Formalizing this optimality is part of follow-up work.

\begin{table}
\centering
 \begin{tabular}{c c} 
 \toprule
 \textbf{Function}  & \textbf{Description}  \\ [0ex]
 \midrule 
 \Train{$i,w$} & Updated model after training $w$ incrementally on client $i$'s data \\ 
 \Thresh{$b$} & Transmission probability based on bid $b$ \\
 \Aggr{$W$} & Aggregated model of models $w \in W$ \\ 
 \Punish{$d$} & (Monetary) punishment based on deviation $d$\\
 \Score{$i,w$} & Score of model $w$ by client $i$ \\
 \bottomrule
\end{tabular}
\caption{Methods used in our algorithm.}\label{tab:algoparameters}
\end{table}

\begin{algorithm}
\SetAlgoVlined
\DontPrintSemicolon
\Input{ Number of federated rounds $T$, Number of clients $k$}
\textbf{Client Initialization:} Each client $i \in [k]$ initializes models $w_i$ and sends bid $b_i$ to server\\

\For{ $t=0,\dots, T$}{

\ForEach{Client $i$}{
$w_i \leftarrow \Train (i, w_i)$\;
}

$(p, s, \hat{s}) \leftarrow \FedEval (w,b)$ \\ 
Server draws a uniformly random permutation $\pi \colon [k] \to [k]$\;
\ForEach{Client $i$}{
\If{$b_i(\hat{s} - s_i) > b_{\pi(i)} (\hat{s}-s_{\pi(i)})$}{
Client $\hat{\iota}$ transfers model $\hat{w}$ giving score $\hat{s}$ to $i$\;
$t_i\leftarrow b_{\pi(i)} (\hat{s}-s_{\pi(i)}) + p_i$
}
\Else{
$t_i \leftarrow p_i$\;
}}}
\Fn{\FedEval{$w,b$}}{
\textbf{Server Initialization:} $a_{ij} \sim \operatorname{Bernoulli}(\Thresh(b_i)), i \neq j \in [k]$, $a_{ii} =1$, $W_{i\to} \coloneqq \{j \in [k]| a_{ij} =1\}$, $W_{\to i} \coloneqq \{j \in [k]| a_{ji} =1\}$\;
\ForEach{Client $i$}{Send $w_i$ to clients $W_{i\to}$}
\ForEach{Client $j$}{Evaluate $\hat{w}_i \leftarrow $\Aggr{$W_{\to j}$} and send $\hat{w}_i$ to Clients $W_{j \to}$}
\ForEach{Client $j$}{Evaluate $s_i^j \leftarrow$ \Score{$j,w_i$} and $\hat{s}^j_i \leftarrow $\Score{$j,\hat{w}_i$}, $i \in [k]$ and send to Server}
\textbf{Server Evaluation:} $s_i \leftarrow \operatorname{med} ((s_i^j)_{j \in W_{i \to}})$, $i \in [k]$\;
$\hat{s}_i \leftarrow \operatorname{med} ((\hat{s}_i^j)_{j \in W_{i \to}})$, $i \in [k]$\;
$\hat {s} \leftarrow \max_{i \in [k]} \hat{s}_i$\;
$p_j \leftarrow \sum_{i=1}^k $\Punish{$s_i^j - s_i$}$+$ \Punish{$\hat{s}_i^j - \hat{s}_i$}\;
\Return{p,s,$\hat{s}$}
}
\caption{Federated Incentive Payments via Prior-Independent Auctions (FIPIA)}
\label{algo:main}
\end{algorithm}
The mechanism starts with the submission of \emph{bids} by clients, which are reports of their valuations $\theta_i$. These reports might or might not be truthful. For each federated around $T$, clients first train on new data they receive. Then, our federated evaluation presented as a subroutine in \autoref{algo:main} determines incentive payments for truthful evaluation, $p$, a vector of scores for each client's model $s$, the score $\hat s$ of the model with the highest median score among all aggregated models. Then, the server matches each client randomly with another client. The server first computes 
\[
b_{\pi(i)} (\hat{s} - s_{\pi(i)})
\]
from the median scores received in the federated evaluation. This value is assuming $\pi(i)$'s report was truthful ($b_{\pi(i)} = \theta_{\pi(i)}$) the gain in objective for  $\pi(s)$ from being allocated the model. Client $i$ gets the best model if their corresponding value 
\[
b_i (\hat{s} - s_i)
\]
surpasses this value. The value for $\pi(i)$ hence serves as minimum payment for $i$ and incentivizes high bidding. If $i$'s value surpasses $\pi(i)$'s value, the system declares $i$ a winner, $i$ gets the best model $\hat w$ which produces score $\hat s$.

Our federated evaluation function defines for each client a random set of receiver clients. Clients that bid higher are more likely to receive models to evaluate. This design choice takes into account that these are also most likely to receive the best model $\hat{w}$ in the course of \autoref{algo:main}. Clients evaluate all single and the average of all the models they receive. By sending these models to other agents and collecting the median of reported scores, the server obtains accurate estimates of individual and best model quality. Payments for reports deviating from the median incentivize truthful evaluations.

\subsection{Design Considerations}
In this subsection, we discuss design considerations for the methods used in \autoref{algo:main}, compare \autoref{tab:algoparameters}. 

In this paper, we consider a semi-honest trust model in which clients follow the protocol specification, but may attempt to learn honest clients' private information from the models it receives, or may collude with other clients to learn private information. Private information of honest clients can be learned by performing model inversion and membership inference attacks \citep{nasr2018comprehensive, melis2019exploiting, geiping2020inverting}. To make our system compatible with the semi-honest trust model, clients can implement \Train{$i,w$} to produce differentially private models \citep{abadi2016deep, wang2019subsampled, wang2020d2p} to prevent these attacks.

The threshold function $\Thresh{b}$ determines how many models clients see in peer evaluations. While a high value for $\Thresh(b)$ yields more accurate estimates of performance, it also induces incentives for lowering bids for clients that hope to receive a high-quality model as an evaluator, and not as a winner of the auction.

The punishment function incentivizes truthful reporting. In cases with heterogenous clients, choosing small values of \Punish{$d$} might be preferable, as otherwise incentives to participate in the system for clients with small test sets might be lowered.

The aggregation function $\Aggr{W}$ can use arbitrary aggregation routines, be they federated averaging \cite{mcmahan2017communication}, or involve search for the best model among the models to aggregate.
\section{Evaluation}
We conducted experiments to evaluate the performance of our mechanism. All code can be found under \url{https://github.com/Indraos/MultiSidedFederation}. We present results for the MNIST dataset \citep{lecun1998mnist}. Our code also runs experiments for the CIFAR-10 \citep{krizhevsky2009learning} and Fashion-MNIST \citep{xiao2017fashion} datasets. The MNIST and FashionMNIST datasets consists of 70000 28x28 images each. There are 60000 training images and 10000 test images. The images are normalized using the training set mean and variance. For MNIST and FashionMNIST, each client trained a simple network with two convolution layers followed by two fully connected layers. The CIFAR-10 dataset consists of 60000 32x32 colour images in 10 classes, with 6000 images per class. There are 50000 training images and 10000 test images. For MNIST and FashionMNIST, each client trained a simple network with two convolution layers followed by three fully connected layers.

Our experiments uses $k=3$ clients, with valuations $0.1$, $0.5$ and $0.6$. Other valuations qualitatively do not change our results. The clients plit the data in an i.i.d. fashion. Training dataset sizes are $50\%$, $40\%$ and $10\%$ of the dataset. To highlight incentive and adaptivity properties of our system without peer evaluation, we set $\operatorname{Punish}(d) =0$ and $\operatorname{Thresh} (b) = 0$.

\begin{figure}
\centering
\begin{subfigure}[c]{.45\textwidth}
\captionsetup{justification=centering,margin=0.3cm}
\includegraphics[width=\textwidth]{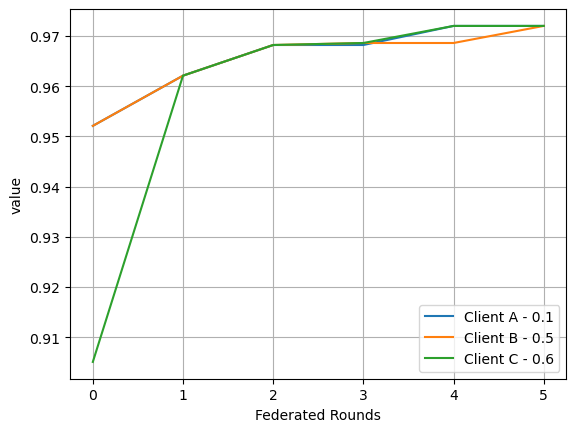}
\caption{Accuracy trajectories for clients in the MNIST setup.}
\label{fig:valuesmnist}
\end{subfigure}  
\begin{subfigure}[c]{.45\textwidth}
\captionsetup{justification=centering,margin=0.cm}
\includegraphics[width=\textwidth]{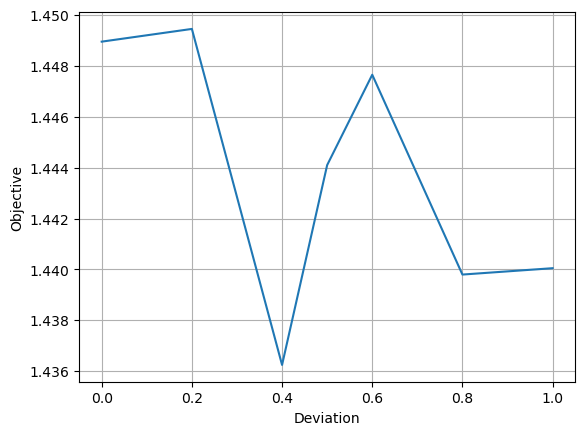}
\caption{Client objective value when bidding a report unequal to valuation $0.5$ in the MNIST setup.}
\label{fig:incentivesmnist}
\end{subfigure}  
\begin{subfigure}[c]{.45\textwidth}
\captionsetup{justification=centering,margin=0.3cm}
\includegraphics[width=\textwidth]{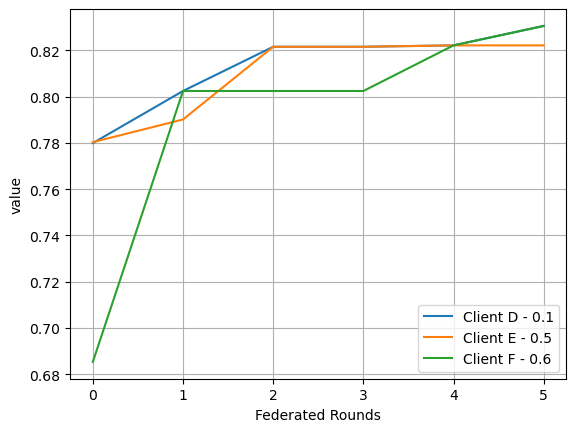}
\caption{Accuracy trajectories for clients in the FashionMNIST setup.}
\label{fig:valuesfashion}
\end{subfigure}  
\begin{subfigure}[c]{.45\textwidth}
\captionsetup{justification=centering,margin=0.3cm}
\includegraphics[width=\textwidth]{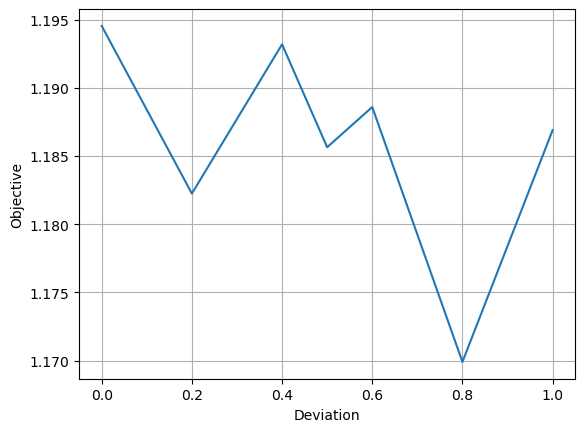}
\caption{Client objective value when bidding a report unequal to valuation $0.5$ in the  FashionMNIST setup.}
\label{fig:incentivesfashion}
\end{subfigure}  
\begin{subfigure}[c]{.45\textwidth}
\captionsetup{justification=centering,margin=0.3cm}
\includegraphics[width=\textwidth]{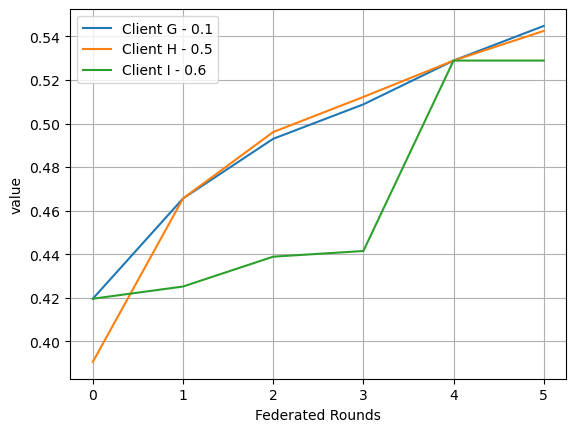}
\caption{Accuracy trajectories for clients in the CIFAR-10 setup.}
\label{fig:valuescifar}
\end{subfigure}  
\begin{subfigure}[c]{.45\textwidth}
\captionsetup{justification=centering,margin=0.3cm}
\includegraphics[width=\textwidth]{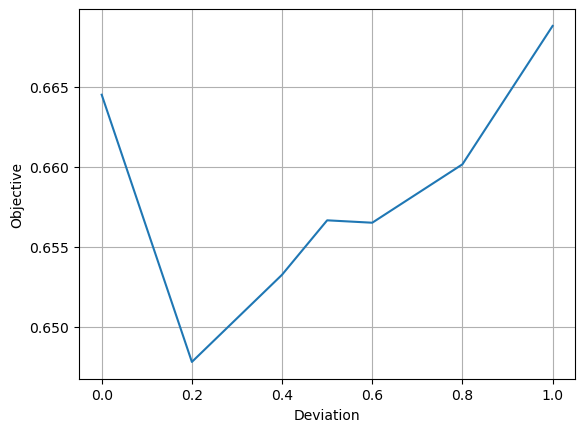}
\caption{Client objective value when bidding a report unequal to valuation $0.5$ in the  CIFAR-10 setup.}
\label{fig:incentivescifar}
\end{subfigure} 
\caption{Experiments for MNIST (top row), FashionMNIST (middle row) and CIFAR-10 (bottom row). The left column shows the values in different clients in our first experiment. The figures in the right column shows for an agent with valuation 0.5 the objective function when submitting bids different from its valuation.}
\label{fig:experiments}
\end{figure}
\subsection{The Effects of Bids and Dataset Sizes}
Our first experiment shows the histories of model qualities for clients with different dataset sizes and bids. We assume that all clients bid their valuation truthfully. \autoref{fig:experiments} represents the accuracies of different clients for the first 6 FL rounds. We note that how the quality of models for different clients adapts in our model according to the bids placed. In particular, the FashionMNIST case shows that the high-bidding model ends up with a higher-quality model than the other clients despite a much smaller dataset, through a combination of being declared winner in the second-to-last round and last-round training.

\subsection{Profitability of Overbidding}
Next, we study the incentive properties of our system, compare the right column of \autoref{fig:experiments}. We calculate the profit from bidding a value different from the valuation $\theta_i$. For our exposition, we restrict to a setting with one client that has a fixed valuation $\theta_i = 0.5$ and the two other agents have valuations $0.1$ and $0.6$. We calculate the values \eqref{eq:util} for client $i$ when it bids $b_i \in \{0,0.2, 0.4, 0.5, 0.6, 0.8, 1\}$. We observe that some deviations are profitable, but that bidding close to truthfully is a or close to a local maximum of the objectives for the deviating agent.
\section{Conclusion}
In this paper, we introduce a system to secure funds from clients participating in FL. Our system is adaptive as it gives models of different qualities to clients that have different valuations. We investigated incentive properties of the system and find some instances of profitability of deviations.

Our work complements existing incentive designs for reimbursing model contributors in FL. In future work, we plan to explore the integration of these systems into an incentive system for the demand \emph{and} supply side of FL and to give theoretical revenue guarantees.
\bibliography{references}

\begin{thebibliography}{23}
\providecommand{\natexlab}[1]{#1}
\providecommand{\url}[1]{\texttt{#1}}
\expandafter\ifx\csname urlstyle\endcsname\relax
  \providecommand{\doi}[1]{doi: #1}\else
  \providecommand{\doi}{doi: \begingroup \urlstyle{rm}\Url}\fi

\bibitem[Abadi et~al.(2016)Abadi, Chu, Goodfellow, McMahan, Mironov, Talwar,
  and Zhang]{abadi2016deep}
Martin Abadi, Andy Chu, Ian Goodfellow, H~Brendan McMahan, Ilya Mironov, Kunal
  Talwar, and Li~Zhang.
\newblock Deep learning with differential privacy.
\newblock In \emph{Proceedings of the 2016 ACM SIGSAC conference on computer
  and communications security}, pp.\  308--318, 2016.

\bibitem[Beaver(1991)]{beaver1991foundations}
Donald Beaver.
\newblock Foundations of secure interactive computing.
\newblock In \emph{Annual International Cryptology Conference}, pp.\  377--391.
  Springer, 1991.

\bibitem[Bonawitz et~al.(2019)Bonawitz, Eichner, Grieskamp, Huba, Ingerman,
  Ivanov, Kiddon, Kone{\v{c}}n{\`y}, Mazzocchi, McMahan,
  et~al.]{bonawitz2019towards}
Keith Bonawitz, Hubert Eichner, Wolfgang Grieskamp, Dzmitry Huba, Alex
  Ingerman, Vladimir Ivanov, Chloe Kiddon, Jakub Kone{\v{c}}n{\`y}, Stefano
  Mazzocchi, H~Brendan McMahan, et~al.
\newblock Towards federated learning at scale: System design.
\newblock \emph{arXiv preprint arXiv:1902.01046}, 2019.

\bibitem[Chen et~al.(2020)Chen, Liu, Shen, Shen, and Tang]{Chen}
Mengjing Chen, Yang Liu, Weiran Shen, Yiheng Shen, and Pingzhong Tang.
\newblock {Mechanism Design for Multi-Party Machine Learning}.
\newblock 2020.

\bibitem[Clarke(1971)]{clarke1971multipart}
Edward~H Clarke.
\newblock Multipart pricing of public goods.
\newblock \emph{Public choice}, pp.\  17--33, 1971.

\bibitem[Feldman \& Chuang(2005)Feldman and
  Chuang]{Feldman2005OvercomingSystems}
Michal Feldman and John Chuang.
\newblock {Overcoming Free-Riding Behavior in Peer-to-Peer Systems}.
\newblock \emph{ACM SIGecom Exchanges}, 5\penalty0 (4):\penalty0 41--50, 2005.

\bibitem[Geiping et~al.(2020)Geiping, Bauermeister, Dr{\"o}ge, and
  Moeller]{geiping2020inverting}
Jonas Geiping, Hartmut Bauermeister, Hannah Dr{\"o}ge, and Michael Moeller.
\newblock Inverting gradients--how easy is it to break privacy in federated
  learning?
\newblock \emph{arXiv preprint arXiv:2003.14053}, 2020.

\bibitem[Goldberg \& Hartline(2001)Goldberg and
  Hartline]{goldberg2001competitive}
Andrew~V Goldberg and Jason~D Hartline.
\newblock Competitive auctions for multiple digital goods.
\newblock In \emph{European Symposium on Algorithms}, pp.\  416--427. Springer,
  2001.

\bibitem[Hartline(2013)]{hartline2013mechanism}
Jason~D Hartline.
\newblock Mechanism design and approximation.
\newblock \emph{Book draft. October}, 122, 2013.

\bibitem[Hsu et~al.(2019)Hsu, Qi, and Brown]{hsu2019measuring}
Tzu-Ming~Harry Hsu, Hang Qi, and Matthew Brown.
\newblock Measuring the effects of non-identical data distribution for
  federated visual classification.
\newblock \emph{arXiv preprint arXiv:1909.06335}, 2019.

\bibitem[Jiao et~al.(2019)Jiao, Wang, Niyato, Lin, and Kim]{Jiao2019}
Yutao Jiao, Ping Wang, Dusit Niyato, Bin Lin, and Dong~In Kim.
\newblock {Toward an automated auction framework for wireless federated
  learning services market}.
\newblock \emph{arXiv}, pp.\  1--14, 2019.
\newblock ISSN 23318422.
\newblock \doi{10.1109/tmc.2020.2994639}.

\bibitem[Krizhevsky et~al.(2009)]{krizhevsky2009learning}
Alex Krizhevsky et~al.
\newblock Learning multiple layers of features from tiny images.
\newblock 2009.

\bibitem[LeCun et~al.(1998)LeCun, Cortes, and Burges]{lecun1998mnist}
Yann LeCun, Corinna Cortes, and Christopher~JC Burges.
\newblock The mnist database of handwritten digits, 1998.
\newblock \emph{URL http://yann. lecun. com/exdb/mnist}, 10:\penalty0 34, 1998.

\bibitem[Li et~al.(2019)Li, Sanjabi, and Smith]{Li2019}
Tian Li, Maziar Sanjabi, and Virginia Smith.
\newblock {Fair resource allocation in federated learning}.
\newblock \emph{arXiv}, pp.\  1--27, 2019.
\newblock ISSN 23318422.

\bibitem[Mas-Colell et~al.(1995)Mas-Colell, Whinston, Green,
  et~al.]{mas1995microeconomic}
Andreu Mas-Colell, Michael~Dennis Whinston, Jerry~R Green, et~al.
\newblock \emph{Microeconomic theory}, volume~1.
\newblock Oxford university press New York, 1995.

\bibitem[McMahan et~al.(2017)McMahan, Moore, Ramage, Hampson, and
  y~Arcas]{mcmahan2017communication}
Brendan McMahan, Eider Moore, Daniel Ramage, Seth Hampson, and Blaise~Aguera
  y~Arcas.
\newblock Communication-efficient learning of deep networks from decentralized
  data.
\newblock In \emph{Artificial Intelligence and Statistics}, pp.\  1273--1282.
  PMLR, 2017.

\bibitem[Melis et~al.(2019)Melis, Song, De~Cristofaro, and
  Shmatikov]{melis2019exploiting}
Luca Melis, Congzheng Song, Emiliano De~Cristofaro, and Vitaly Shmatikov.
\newblock Exploiting unintended feature leakage in collaborative learning.
\newblock In \emph{2019 IEEE Symposium on Security and Privacy (SP)}, pp.\
  691--706. IEEE, 2019.

\bibitem[Mugunthan et~al.(2020)Mugunthan, Rahman, and Kagal]{Mugunthan2020a}
Vaikkunth Mugunthan, Ravi Rahman, and Lalana Kagal.
\newblock {BlockFLow: An Accountable and Privacy-Preserving Solution for
  Federated Learning}.
\newblock \emph{arXiv}, 2020.

\bibitem[Nasr et~al.(2018)Nasr, Shokri, and Houmansadr]{nasr2018comprehensive}
Milad Nasr, Reza Shokri, and Amir Houmansadr.
\newblock Comprehensive privacy analysis of deep learning: Stand-alone and
  federated learning under passive and active white-box inference attacks.
\newblock \emph{arXiv preprint arXiv:1812.00910}, 2018.

\bibitem[Wang et~al.(2020)Wang, Jia, and Song]{wang2020d2p}
L~Wang, R~Jia, and D~Song.
\newblock D2p-fed: Differentially private federated learning with efficient
  communication.
\newblock \emph{arxiv. org/pdf/2006.13039}, 2020.

\bibitem[Wang et~al.(2019)Wang, Balle, and Kasiviswanathan]{wang2019subsampled}
Yu-Xiang Wang, Borja Balle, and Shiva~Prasad Kasiviswanathan.
\newblock Subsampled r{\'e}nyi differential privacy and analytical moments
  accountant.
\newblock In \emph{The 22nd International Conference on Artificial Intelligence
  and Statistics}, pp.\  1226--1235. PMLR, 2019.

\bibitem[Xiao et~al.(2017)Xiao, Rasul, and Vollgraf]{xiao2017fashion}
Han Xiao, Kashif Rasul, and Roland Vollgraf.
\newblock Fashion-mnist: a novel image dataset for benchmarking machine
  learning algorithms.
\newblock \emph{arXiv preprint arXiv:1708.07747}, 2017.

\bibitem[Yu et~al.(2020)Yu, Liu, Liu, Chen, Cong, Weng, Niyato, and
  Yang]{yu2020fairness}
Han Yu, Zelei Liu, Yang Liu, Tianjian Chen, Mingshu Cong, Xi~Weng, Dusit
  Niyato, and Qiang Yang.
\newblock A fairness-aware incentive scheme for federated learning.
\newblock In \emph{Proceedings of the AAAI/ACM Conference on AI, Ethics, and
  Society}, pp.\  393--399, 2020.

\end{thebibliography}
\bibliographystyle{iclr2021_conference}
\end{document}